\documentclass[11pt,a4paper]{article}
\usepackage[hyperref]{acl2020}
\usepackage{times}
\usepackage{latexsym}
\usepackage{microtype}
\usepackage{amsmath}
\usepackage{amsfonts}
\usepackage{algorithm}
\usepackage{booktabs}
\usepackage{adjustbox}
\usepackage{url}
\usepackage{supertabular}
\usepackage{xcolor}
\usepackage{tablefootnote}
\usepackage{xcolor}
\usepackage{lscape}
\usepackage{dirtytalk}

\usepackage{url}

\aclfinalcopy

\title{Intermediate-Task Transfer Learning with Pretrained Models for Natural Language Understanding: When and Why Does It Work?}

\author{Yada Pruksachatkun$^1$\thanks{~~Equal contribution.}\ \ ~~Jason Phang$^1$\footnotemark[1]\ \  ~~Haokun Liu$^1$\footnotemark[1]\ \ ~~Phu Mon Htut$^1$\footnotemark[1]\\ \bf Xiaoyi Zhang$^1$\ \  Richard Yuanzhe Pang$^1$\ \ Clara Vania$^1$\ \ Katharina Kann$^2$\\ \bf Samuel R. Bowman$^1$\\
$^1$New York University \\
$^2$University of Colorado Boulder\\
\texttt{\{yp913,bowman\}@nyu.edu}
}

\date{}

\begin{document}
\maketitle
\begin{abstract}

While pretrained models such as BERT have shown large gains across natural language understanding tasks, their performance can be improved by further training the model on a data-rich \textit{intermediate task}, before fine-tuning it on a target task.
However, it is still poorly understood when and why intermediate-task training is beneficial for a given target task.
To investigate this, we perform a large-scale study on the pretrained RoBERTa 
model  with 110 intermediate--target task combinations. 
We further evaluate all trained models with 25 \textit{probing tasks} meant to reveal the specific skills that drive transfer.
We observe that intermediate tasks requiring high-level inference and reasoning abilities tend to work best.
We also  observe that target task performance is strongly correlated with higher-level abilities such as coreference resolution.
However, we fail to observe more granular correlations between probing and target task performance, highlighting the need for further work on broad-coverage probing benchmarks. We also observe evidence that the forgetting of knowledge learned during pretraining may limit our analysis, highlighting the need for further work on transfer learning \textit{methods} in these settings.
\end{abstract}

\section{Introduction}
Unsupervised pretraining---e.g., BERT \citep{devlin-etal-2019-bert} or RoBERTa \citep{liu2019roberta}---has recently pushed the state of the art on many natural language understanding tasks. 
One method of further improving pretrained models that has been shown to be broadly helpful is to first fine-tune a pretrained model on an \textit{intermediate} task, before fine-tuning again on the target task of interest \citep{Phang2018SentenceEO,wang-etal-2019-tell,clark-etal-2019-boolq,sap-etal-2019-social}, also referred to as STILTs. However, this approach does not always improve target task performance, and it is unclear under what conditions it does.

\begin{figure}
\centering
  \includegraphics[width=\columnwidth]{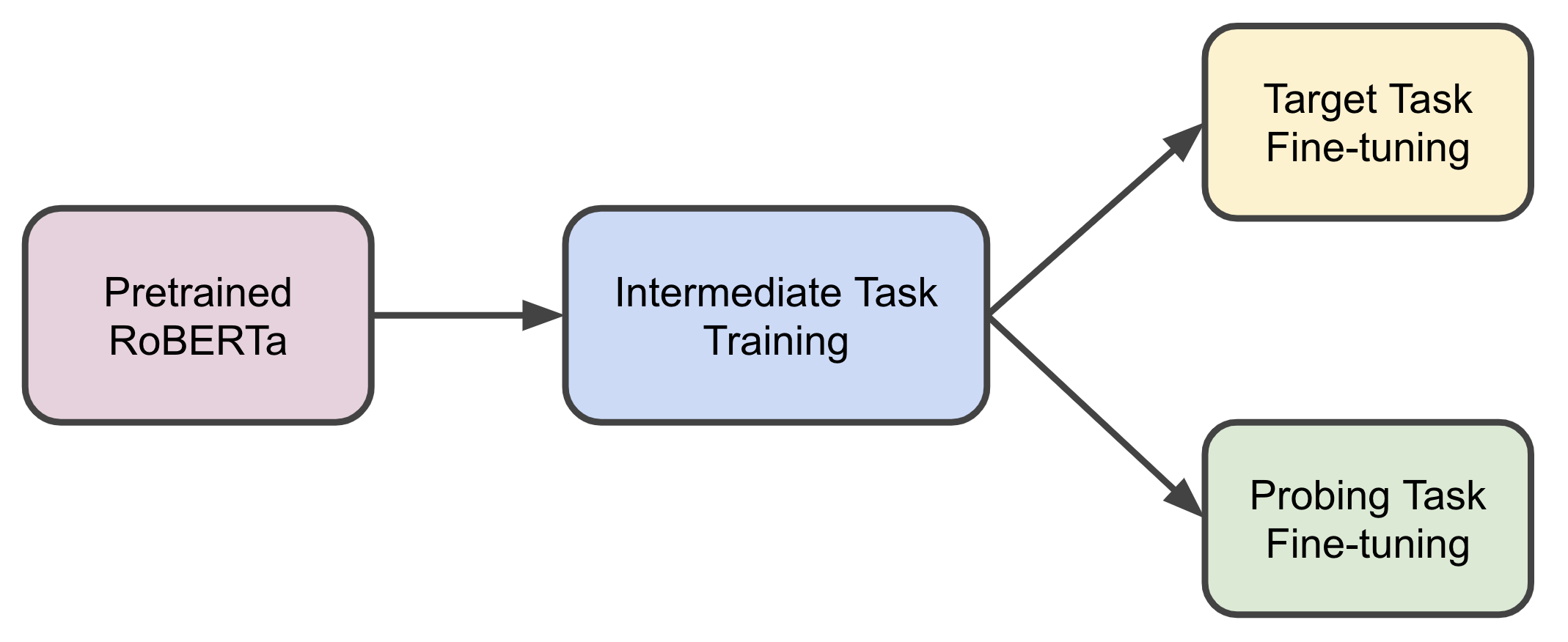}
  \caption{Our experimental pipeline with intermediate-task transfer learning and subsequent fine-tuning on target and probing tasks.}
  \label{fig:exp-pipeline}
\end{figure}

This paper offers a large-scale empirical study aimed at addressing this open question. We perform a broad survey of intermediate and target task pairs, following an experimental pipeline similar to \citet{Phang2018SentenceEO} and \citet{wang-etal-2019-tell}. This differs from previous work in that we use a larger and more diverse set of intermediate and target tasks, introduce additional analysis-oriented probing tasks, and use a better-performing base model RoBERTa \citep{liu2019roberta}.
We aim to answer the following specific questions:
\begin{itemize}
    \item What kind of tasks tend to make good intermediate tasks across a wide variety of target tasks?
    \item Which linguistic skills does a model learn from intermediate-task training?
    \item Which skills learned from intermediate tasks help the model succeed on which target tasks?
\end{itemize}
The first question is the most straightforward: it can be answered by a sufficiently exhaustive search over possible intermediate--target task pairs. The second and third questions address the \textit{why} rather than the \textit{when}, and differ in a crucial detail: A model might learn skills by training on an intermediate task, but those skills might not help it to succeed on a target task. 

Our search for intermediate tasks focuses on natural language understanding tasks in English.
In particular, we run our experiments on 11 intermediate tasks and 10 target tasks, which results in a total of 110 intermediate--target task pairs. We use 25 \textit{probing tasks}---tasks that each target a narrowly defined model behavior or linguistic phenomenon---to shed light on which skills are learned from each intermediate task. 

Our findings include the following: 
(i) Natural language inference tasks as well as QA tasks which involve commonsense reasoning are generally useful as intermediate tasks. 
(ii) SocialIQA and QQP as intermediate tasks are not helpful as a means to teach the skills captured by our probing tasks, while finetuning first on  MNLI and CosmosQA result in an increase in all skills. 
(iii) While a model's ability to learn skills relating to input-noising correlate with target task performance, low-level skills such as knowledge of a sentence's raw content preservation skills and ability to detect various attributes of input sentences such as tense of main verb and sentence length are less correlated with target task performance. This suggests that a model's ability to do well on the masked language modelling (MLM) task is important for downstream performance. Furthermore, we conjecture that a portion of our analysis is affected by catastrophic forgetting of knowledge learned during pretraining.

\section{Methods}

\subsection{Experimental Pipeline}
Our experimental pipeline (Figure \ref{fig:exp-pipeline}) consists of two steps, starting with a pretrained model: \textit{intermediate-task training}, and \textit{fine-tuning} on a \textit{target} or \textit{probing} task. 

\paragraph{Intermediate Task Training}

We fine-tune RoBERTa on each intermediate task. The training procedure follows the standard procedure of fine-tuning a pretrained model on a target task, as described in \citet{devlin-etal-2019-bert}. We opt for single intermediate-task training as opposed to multi-task training \citep[cf.][]{liu-etal-2019-multi} to isolate the effect of skills learned from individual intermediate tasks.

\paragraph{Target and Probing Task Fine-Tuning}
After intermediate-task training, we fine-tune our models on each target and probing task individually. Target tasks are tasks of interest to the general community, spanning various facets of natural language, domains, and sources. 
Probing tasks, while potentially similar in data source to target tasks such as with CoLA, are designed to isolate the presence of particular linguistic capabilities or skills.  For instance, solving the target task BoolQ \citep{clark-etal-2019-boolq} may require various skills including coreference and commonsense reasoning, while probing tasks like the SentEval probing suite \citep{conneau-etal-2018-cram} target specific syntactic and metadata-level phenomena such as subject-verb agreement and sentence length detection.
\makeatletter
\newcommand\footnoteref[1]{\protected@xdef\@thefnmark{\ref{#1}}\@footnotemark}
\makeatother

\newcommand*\rot{\rotatebox{90}}

 \begin{table*}[t]
  \setlength{\tabcolsep}{2pt}
  \small
  \centering
  \begin{tabular}{llrrlll}
    \toprule
    & \textbf{Name} & \textbf{$|$Train$|$}  & \textbf{$|$Dev$|$}  & \textbf{task} & \textbf{metrics} & \textbf{genre/source} \\ 
    \midrule

    & CommonsenseQA & 9,741 & 1,221 & question answering & acc. & ConceptNet \\  
    & SciTail & 23,596 & 1,304 & natural language inference & acc. & science exams  \\
    & Cosmos QA  & 25,588 & 3,000 & question answering & acc. & blogs \\
    & SocialIQA & 33,410 & 1,954 & question answering & acc. & crowdsourcing \\
    & CCG & 38,015 & 5,484 & tagging & acc. & Wall Street Journal\\
    & HellaSwag & 39,905 & 10,042 & \raggedright sentence completion & acc. & video captions \& Wikihow \\
    & QA-SRL & 44,837 & 7,895 & question answering & F1/EM & Wikipedia \\
    & SST-2 & 67,349 & 872 & sentiment classification & acc. & movie reviews\\
        & QAMR & 73,561 & 27,535 & question answering  & F1/EM & Wikipedia  \\
    \rot{\rlap{\textbf{Intermediate Tasks}}}
    & QQP & 363,846 & 40,430 & paraphrase detection & acc./F1 & Quora questions\\
    & MNLI & 392,702 & 20,000 & natural language inference & acc. & fiction, letters, telephone speech\\
    \midrule
    & CB  & 250 & 57 & natural language inference & acc./F1 & Wall Street Journal, fiction, dialogue \\
    & COPA & 400 & 100 & question answering & acc. & blogs, photography encyclopedia \\
    & WSC & 554 & 104 & coreference resolution & acc. & hand-crafted  \\
    & RTE & 2,490 & 278 & natural language inference  & acc. & news, Wikipedia\\
    & MultiRC  & 5,100 & 953 & question answering & F1$_{\alpha}$/EM & crowd-sourced\\
    & WiC  & 5,428 & 638 & word sense disambiguation & acc. & WordNet, VerbNet, Wiktionary  \\
    & BoolQ & 9,427 & 3,270 & question answering & acc. & Google queries, Wikipedia \\
    \rot{\rlap{\textbf{Target Tasks}}}
    & CommonsenseQA & 9,741 & 1,221 & question answering  & acc. & ConceptNet \\
    & Cosmos QA & 25,588 & 3,000 & question answering & acc. & blogs \\
    & ReCoRD & 100,730 & 10,000 & question answering & F1/EM & news (CNN, Daily Mail)\\
    \bottomrule
  \end{tabular}
  \caption{Overview of the intermediate tasks ({top}) and target tasks ({bottom}) in our experiments. EM is short for Exact Match. 
 The F1 metrics for MultiRC is calculated over all answer-options. }
  \label{tab:stats}
\end{table*}
\subsection{Tasks}
\label{sec:tasks}

Table \ref{tab:stats} presents an overview of the intermediate and target tasks. 

\subsubsection{Intermediate Tasks}
\label{sec:interm-task}
We curate a diverse set of tasks that either represent an especially large annotation effort or that have been shown to yield positive transfer in prior work. The resulting set of tasks cover question answering, commonsense reasoning, and natural language inference. 
\paragraph{QAMR}
The Question--Answer Meaning
Representations dataset \citep{qamr} is a crowdsourced QA task consisting of question--answer pairs that correspond to predicate--argument relationships. It is derived from Wikinews and Wikipedia sentences. For example, if the sentence is \say{\textit{Ada Lovelace was a computer scientist.}}, a potential question is \say{\textit{What is Ada's last name?}}, with the answer being \say{\textit{Lovelace.}}

\paragraph{CommonsenseQA}
CommonsenseQA \cite{commonsenseqa} is a multiple-choice QA task derived from ConceptNet \citep{conceptnet} with the help of crowdworkers, that is designed to test a range of commonsense knowledge.

\paragraph{SciTail} SciTail \cite{scitail} is a textual entailment task built from  multiple-choice science questions from 4th grade and 8th grade exams, as well as crowdsourced questions \citep{welbl}. The task is to determine whether a hypothesis, which is constructed from a science question and its corresponding answer, is entailed or not (neutral) by the premise.

\paragraph{Cosmos QA}
Cosmos QA is a task for a commonsense-based reading comprehension task formulated as multiple-choice questions \citep{cosmos}. The questions concern the causes or effects of events that require reasoning not only based on the exact text spans in the context, but also wide-range abstractive commonsense reasoning. It differs from CommonsenseQA in that it focuses on causal and deductive commensense reasoning and that it requires reading comprehension over an auxiliary passage, rather than simply answering a freestanding question.

\paragraph{SocialIQA} SocialIQA \citep{sap-etal-2019-social} is a task for multiple choice QA. It tests for reasoning surrounding emotional and social intelligence in everyday situations.

\paragraph{CCG}
CCGbank \citep{ccg} is a task that is a translation of the Penn Treebank into a corpus of Combinatory Categorial Grammar (CCG) derivations. We use the CCG supertagging task, which is the task of assigning tags to individual word tokens that jointly determine the parse of the sentence.

\paragraph{HellaSwag}
HellaSwag \citep{hellaswag} is a commonsense reasoning task that tests a model's ability to choose the most plausible continuation of a story. It is  built using adversarial filtering \citep{zellers-etal-2018-swag} with BERT to create challenging  negative examples.

\paragraph{QA-SRL} 
The question-answer driven semantic role labeling dataset \citep[QA-SRL;][]{he-etal-2015-question} for a QA task that is derived from a semantic role labeling task. Each example, which consists of  a set of questions and answers, corresponds to a predicate-argument relationship in the sentence it is derived from. Unlike QAMR, which focuses on all words in the sentence, QA-SRL is specifically focused on verbs.

\paragraph{SST-2}
The Stanford sentiment treebank \citep{sst2} is a sentiment classification task based on movie reviews. We use the binary sentence classification version of the task.

\paragraph{QQP} The Quora Question Pairs dataset\footnote{http://data.quora.com/First-Quora-DatasetRelease-Question-Pairs \label{qqp-source}} is constructed based on questions posted on the community question-answering website Quora. The task is to determine if two questions are semantically equivalent.

\paragraph{MNLI} The Multi-Genre Natural Language Inference dataset \citep{mnli} is a crowdsourced collection of sentence pairs with textual entailment annotations across a variety of genres. 

\subsubsection{Target Tasks}
\label{sec:target-task}

We use ten target tasks, eight of which are drawn from the SuperGLUE benchmark \citep{wang2019superglue}. The tasks in the SuperGLUE benchmark cover question answering, entailment, word sense disambiguation, and coreference resolution and have been shown to be easy for humans but difficult for models like BERT. Although we offer a brief description of the tasks below, we refer readers to the SuperGLUE paper for a more detailed description of the tasks.

\textbf{CommitmentBank} \citep[\textbf{CB};][]{de2019commitmentbank} is a three-class entailment task that consists of texts and an embedded clause that appears in each text, in which models must determine whether that embedded clause is entailed by the text. \textbf{Choice of Plausible Alternatives} \citep[\textbf{COPA};][]{roemmele2011choice} is a classification task that consists of premises and a question that asks for the cause or effect of each premise, in which models must correctly pick between two possible choices. \textbf{Winograd Schema Challenge} \citep[\textbf{WSC};][]{wsc} is a sentence-level commonsense reasoning task that consists of texts, a pronoun from each text, and a list of possible noun phrases from each text. The dataset has been designed such that world knowledge is required to determine which of the possible noun phrases is the correct referent to the pronoun. We use the SuperGLUE binary classification cast of the task, where each example consists of a text, a pronoun, and a noun phrase from the text, which models must classify as being coreferent to the pronoun or not. \textbf{Recognizing Textual Entailment} \citep[\textbf{RTE};][et seq]{dagan2005pascal} is a textual entailment task.
\textbf{Multi-Sentence Reading Comprehension} \citep[\textbf{MultiRC};][]{multirc} is a multi-hop QA task that consists of  paragraphs, a question on each paragraph, and a list of possible answers, in which models must distinguish which of the possible answers are true and which are false. 
\textbf{Word-in-Context} \citep[\textbf{WiC};][]{wic} is a binary classification word sense disambiguation task. Examples consist of two text snippets, with a polysemous word that
appears in both.  Models must determine whether the same sense of the word is used in both contexts.
 \textbf{BoolQ} \citep{clark-etal-2019-boolq} is a QA task that consists of passages and a yes/no question associated with each passage.
\textbf{Reading Comprehension with Commonsense Reasoning} \citep[\textbf{ReCoRD};][]{record}  is a multiple-choice QA task that consists of news articles. For each article, models are given a question about each article with one entity masked out and a list of possible entities from the article, and the goal is to  correctly identify the masked entity out of the list. 

Additionally, we use \textbf{CommonsenseQA} and \textbf{Cosmos QA} as target tasks, due to their unique combination of small dataset size and high level of difficulty for high-performing models like BERT from our set of intermediate tasks.
\subsubsection{Probing Tasks}
\label{sec:probing-task}

We use  well-established datasets for our probing tasks, including the edge-probing suite from \citet{tenney2018what}, function word oriented tasks from \citet{kim-etal-2019-probing},  and sentence-level probing datasets \citep[SentEval;][]{conneau-etal-2018-cram}. 

\paragraph{Acceptability Judgment Tasks} 

This set of binary classifications tasks was designed to investigate if a model can judge the grammatical acceptability of a sentence. We use the following five datasets: \textbf{AJ-CoLA} is a task that tests for a model's understanding of general grammaticality using the Corpus of Linguistic Acceptability (\text{CoLA}) \citep{warstadt2018neural}, which is drawn from 22 theoretical linguistics publications. The other tasks concern the behaviors of specific classes of function words, using the dataset by \citet{kim-etal-2019-probing}: \textbf{AJ-WH} is a task that tests a model's ability to detect if a wh-word in a sentence has been swapped with another wh-word, which tests a model's ability to identify the antecedent associated with the wh-word. \textbf{AJ-Def} is a task that tests a model's ability to detect if the definite/indefinite articles in a given sentence have been swapped. \textbf{AJ-Coord} is a task that tests a model's ability to detect if a coordinating conjunction has been swapped, which tests a model's ability to understand how ideas in the various clauses relate to each other. \textbf{AJ-EOS} is a task that tests a model's ability to identify  grammatical sentences without indicators such as punctuation marks and capitalization, and consists of grammatical text that are removed of punctuation.

\paragraph{Edge-Probing Tasks}

The edge probing (EP) tasks are a set of core NLP labeling tasks, collected by \citet{tenney2018what} and cast into Boolean classification. These tasks focus on the syntactic and semantic relations between spans in a sentence. The first five tasks use the OntoNotes corpus \citep{Hovy:2006:O9S:1614049.1614064}: \textbf{Part-of-Speech tagging} (\textbf{EP-POS}) is a task that tests a model's ability to  predict the syntactic category (noun, verb, adjective, etc.) for each word in the sentence. \textbf{Named entity recognition} (\textbf{EP-NER}) is  task that tests a model's ability to predict the category of an entity in a {given} span. \textbf{Semantic Role Labeling} (\textbf{EP-SRL}) is a task that tests a model's ability to assign a label to a given span of words that indicates its semantic role (agent, goal, etc.) in the sentence. \textbf{Coreference} (\textbf{EP-Coref}) is a task that tests a model's ability to classify if two spans of tokens refer to the same entity/event. 

The other datasets can be broken down into both syntactic and semantic probing tasks. \textbf{Constituent labeling} (\textbf{EP-Const}) is a task that tests a model's ability to classify a non-terminal label for a span of tokens (e.g., noun phrase, verb phrase, etc.).
\textbf{Dependency labeling} (\textbf{EP-UD}) is a task that tests a model on the functional relationship of one token relative to another. We use the English Web Treebank portion of Universal Dependencies 2.2 release \citep{silveira-etal-2014-gold} for this task. \textbf{Semantic Proto-Role labeling} is a task that tests a model's ability to predict the fine-grained non-exclusive semantic attributes of a given span. Edge probing uses two datasets for SPR: SPR1 (\textbf{EP-SPR1}) \citep{teichert2017semantic}, derived from the Penn Treebank, and SPR2 (\textbf{EP-SPR2}) \citep{rudinger-etal-2018-neural}, derived from the English Web Treebank. \textbf{Relation classification} (\textbf{EP-Rel}) is a task that tests a model's ability to predict the relation between two entities. We use the SemEval 2010 Task 8 dataset \citep{hendrickx-etal-2009-semeval} for this task. For example, the relation between \say{Yeri} and \say{Korea} in \say{Yeri is from Korea} is \textsc{ENTITY-ORIGIN}. The \textbf{Definite Pronoun Resolution} dataset \citep{rahman-ng-2012-resolving} (\textbf{EP-DPR}) is a task that tests a model's ability to handle coreference, and differs from OntoNotes in that it focuses on difficult cases of definite pronouns.

\paragraph{SentEval Tasks}

The SentEval probing tasks (SE) \citep{conneau-etal-2018-cram} are cast in the form of single-sentence classification.  
\textbf{Sentence Length} (\textbf{SE-SentLen}) is a task that tests a model's ability to classify the length of a sentence. \textbf{Word Content} (\textbf{SE-WC}) is a task that tests a model's ability to identify which of a set of 1,000 potential words appear in a given sentence. \textbf{Tree Depth} (\textbf{SE-TreeDepth}) is a task that tests a model's ability to estimate the maximum depth of the constituency parse tree of the sentence. \textbf{Top Constituents} (\textbf{SE-TopConst}) is a task that tests a model's ability to identify the high-level syntactic structure of the sentence by choosing among 20 constituent sequences (the 19 most common, plus an \textit{other} category). \textbf{Bigram Shift} (\textbf{SE-BShift}) is a task that tests a model's ability to classify if two consecutive tokens in the same sentence have been reordered.
\textbf{Coordination Inversion} (\textbf{SE-CoordInv}) is a task that tests a model's ability to identify if two coordinating clausal conjoints are swapped (\textit{ex:} ``he knew it, and he deserved no answer.''). 
\textbf{Past-Present} (\textbf{SE-Tense}) is a task that tests a model's ability to classify the tense of the main verb of the sentence. \textbf{Subject Number} (\textbf{SE-SubjNum}) and \textbf{Object Number} (\textbf{SE-ObjNum}) are tasks that test a model's ability to classify whether the subject or direct object of the main clause is singular or plural. \textbf{Odd-Man-Out} (\textbf{SE-SOMO}) is a task that tests the model's ability to predict whether a sentence has had one of its content words randomly replaced with another word of the same part of speech.
\begin{figure*}[t]
\centering  
\includegraphics[width=1\textwidth]{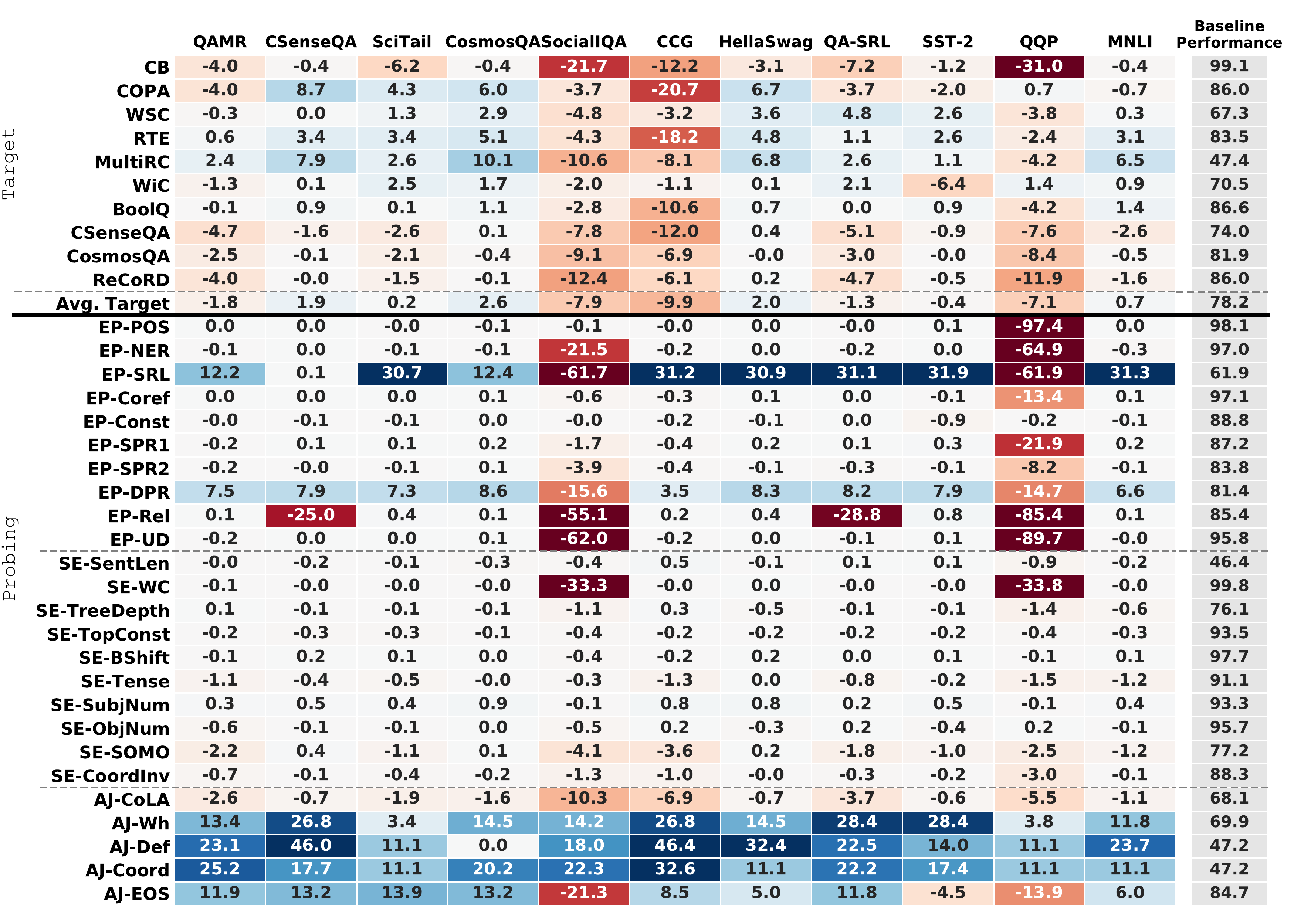}
  \caption{Transfer learning results between intermediate and target/probing tasks. 
  Baselines (rightmost column) are models fine-tuned without intermediate-task training. Each cell shows the difference in performance (delta) between the baseline and model with intermediate-task training. We use the macro-average of each task's metrics as the reported performance. Refer to Table~\ref{tab:stats} for target task metrics.
  }
  \label{fig:transfer1}
\end{figure*} 

\section{Experiments}
\label{sec:exp}

\paragraph{Training and Optimization}

We use the large-scale pretrained model RoBERTa$_{\textrm{Large}}$ 
in all experiments. For each intermediate, target, and probing task, we perform a hyperparameter sweep, 
varying the peak learning rate $ \in \{2\times 10^{-5}, 1\times 10^{-5}, 5\times 10^{-6}, 3\times 10^{-6}\}$ and the dropout rate $ \in \{0.2, 0.1\}$.  After choosing the best learning rate and dropout rate, we apply the best configuration for each task for all runs. For each task, we use the batch size that maximizes GPU usage, and use a maximum sequence length of 256. Aside from these details, we follow the RoBERTa paper for all other training hyperparameters. We use NVIDIA P40 GPUs for our experiments.

A complete pipeline with one intermediate task works as follows: First, we fine-tune RoBERTa on the intermediate task. We then fine-tune copies of the resulting model separately on each of the 10 target tasks and 25 probing tasks and test on their respective validation sets. We run the same pipeline three times for the 11 intermediate tasks, plus a set of baseline runs without intermediate training. This gives us 35$\times$12$\times$3 = 1260 observations.

We train our models using the Adam optimizer \citep{KingmaB14} with linear decay and early stopping. We run training for a maximum of 10 epochs when more than 1,500 training examples are available, and 40 epochs otherwise to ensure models are sufficiently trained on small datasets.
We use the \texttt{jiant} \citep{wang2019jiant} NLP toolkit, based on PyTorch \citep{NEURIPS2019_9015}, Hugging Face Transformers \citep{Wolf2019HuggingFacesTS}, and AllenNLP \citep{Gardner2017AllenNLP}, for all of our experiments.

\section{Results and Analysis}

\label{sec:results}
\subsection{Investigating Transfer Performance}
 Figure \ref{fig:transfer1} shows the differences in target and probing task performances (deltas) between the baselines and models trained with intermediate-task training, each averaged across three restarts. A positive delta indicates successful transfer.

\paragraph{Target Task Performance}
We define good intermediate tasks as ones that lead to positive transfer in target task performance. We observe that tasks that require complex reasoning and inference tend to make good intermediate tasks. These include MNLI and commonsense-oriented tasks such as CommonsenseQA, HellaSWAG, and Cosmos QA (with our poor performance with the similar  SocialIQA serving as a suprising exception). SocialIQA, CCG, and QQP as intermediate tasks lead to negative transfer on all target tasks and the majority of probing tasks. 

We investigate the role of dataset size in the intermediate tasks with downstream task performance by additionally running a set of experiments on varying amounts of data on five intermediate tasks, which is shown in the Appendix. We do not find differences in intermediate-task dataset size to have any substantial consistent impact on downstream target task performance. 

In addition, we find that smaller target tasks such as RTE, BoolQ, MultiRC, WiC, WSC benefit the most from intermediate-task training.\footnote{The deltas for experiments with the same intermediate and target tasks are not 0 as may be expected. This is because we perform both intermediate and target training phases in these cases, with reset optimizer states and stopping criteria in between intermediate and target training.} There are no instances of positive transfer to CommitmentBank, since our baseline model achieves $100\%$ accuracy. 

\paragraph{Probing Task Performance}

Looking at the probing task performance, we find that intermediate-task training affects performance on low-level syntactic probing tasks uniformly across intermediate tasks; we observe little to no improvement for the SentEval probing tasks and higher improvement for acceptability judgment probing tasks, except for AJ-CoLA. This is also consistent with \citet{Phang2018SentenceEO}, who find negative transfer with CoLA in their experiments.

\paragraph{Variation across Intermediate Tasks}
There is variable performance across higher-level syntactic or semantic tasks such as the Edge-Probing and SentEval tasks. SocialIQA and QQP have negative transfer for most of the Edge-Probing tasks, while CosmosQA and QA-SRL see drops in performance only for EP-Rel. While we do see that intermediate-task trained models improve performance on EP-SRL and EP-DPR across the board, there is little to no gain in SentEval probing tasks from any intermediate tasks. Additionally, tasks that increase performance in the most number of probing tasks perform well as intermediate tasks.

\paragraph{Degenerate Runs}

We find that the model may not exceed chance performance in some training runs. This mostly affects the baseline (no intermediate training) runs on the acceptability judgment probing tasks, excluding AJ-CoLA, which all have very small training sets. We include these degenerate runs in our analysis to reflect this phenomenon.  Consistent with \citet{Phang2018SentenceEO}, we find that intermediate-task training reduces the likelihood of degenerate runs, leading to ostensibly positive transfer results on those four acceptability judgment tasks across most intermediate tasks. On the other hand, extremely negative transfer from intermediate-task training can also result in a higher frequency of degenerate runs in downstream tasks, as we observe in the cases of using QQP and SocialIQA as intermediate tasks. We also observe a number of degenerate runs on the EP-SRL task as well as the EP-Rel task. These degenerate runs decrease positive transfer in probing tasks, such as with SocialIQA and QQP probing performance, and also decrease the average amount of positive transfer we see in target task performance. 

\subsection{Correlation Between Probing and Target Task Performance}

\begin{figure*}[t!]
\centering
  \includegraphics[width=1\textwidth]{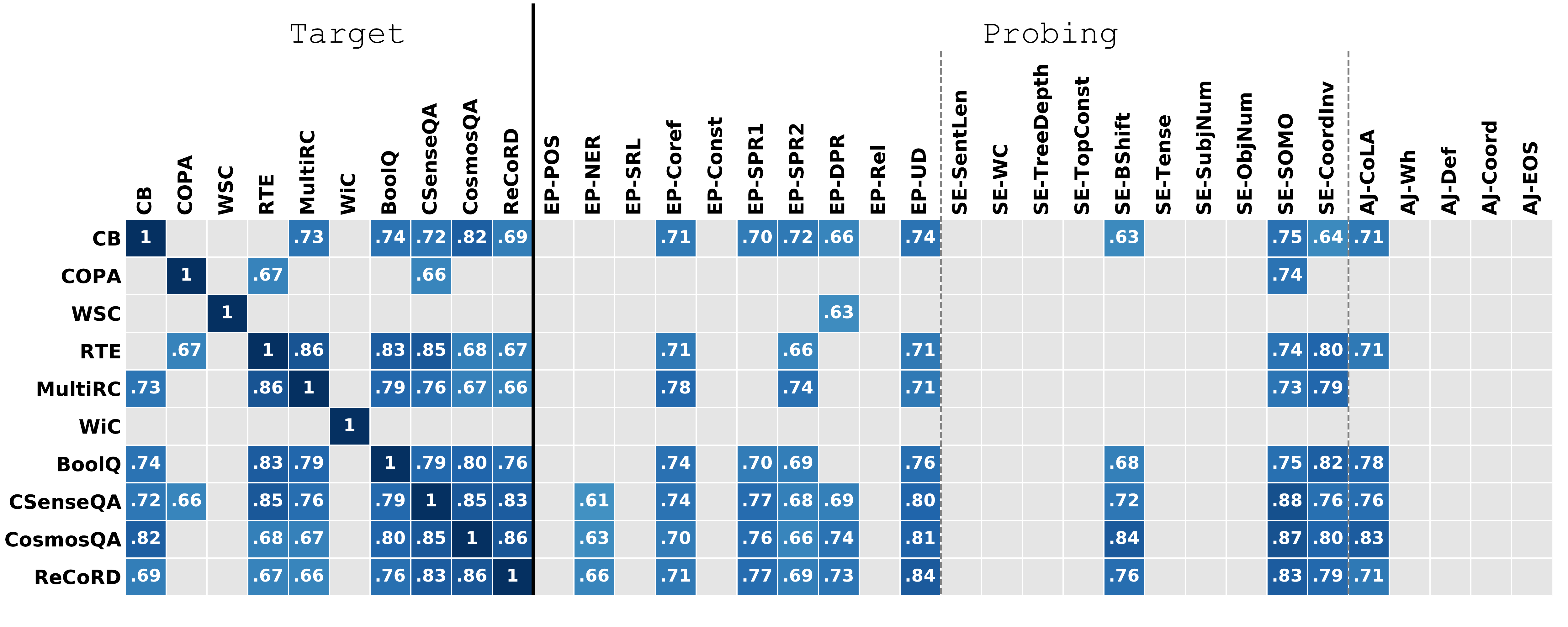}
  \caption{Correlations between probing and target task performances. Each cell contains the Spearman correlation between probing-task and target-task performances across training on different intermediate tasks and random restarts. We test for statistical significance at $p=0.05$ with Holm-Bonferroni correction, and omit the correlations that are not statistically significant.}
  \label{fig:correl_small}
\end{figure*}

Next, we investigate the relationship between target and probing tasks in an attempt to understand \textit{why} certain intermediate-task models perform better on certain target tasks. 

We use probing task performance as an indicator of the acquisition of particular language skills. We compute the Spearman correlation between probing-task and target-task performances across training on different intermediate tasks and multiple restarts, as shown in Figure \ref{fig:correl_small}. We test for statistical significance at $p=0.05$ and apply Holm-Bonferroni correction for multiple testing. We omit correlations that are not statistically significant. We opt for Spearman and not Pearson correlation because of the wide variety of metrics used for the different tasks.\footnote{Full correlation tables across all target and probing tasks with both Spearman and Pearson correlations can be found in the Appendix.}

We find that acceptability judgment probing task performance is generally uncorrelated with the target task performance, except for AJ-CoLA. Similarly, many of the SentEval tasks do not correlate with the target tasks, except for Bigram Shift (SE-BShift), Odd-Man-Out (SE-SOMO) and Coordination Inversion (SE-CoordInv). These three tasks are input noising tasks---tasks where a model has to predict if a given input sentence has been randomly modified---which are, by far, the most similar tasks we study to the masked language modeling task that is used for training RoBERTa. This may explain the strong correlation with the performance of the target tasks. 

We also find that some of these strong correlations, such as with SE-SOMO and SE-CoordInv, are almost entirely driven by variation in the degree of negative transfer, rather than any positive transfer. Intuitively, fine-tuning RoBERTa on an intermediate task can cause the model to forget some of its ability to perform the MLM task. Thus, a future direction for potential improvement for intermediate-task training may be integrating the MLM objective into intermediate-task training or bounding network parameter changes to reduce \textit{catastrophic forgetting} \citep{Kirkpatrick_forgetting,NIPS2019_Forgetting}.

Interestingly, while intermediate tasks such as  SocialIQA, CCG and QQP, which show negative transfer on target tasks, tend to have negative transfer on these three probing tasks, the intermediate tasks with positive transfer, such as CommonsenseQA tasks and MNLI, do not appear to adversely affect the performance on these probing tasks. This asymmetric impact may indicate that, beyond the similarity of intermediate and target tasks, avoiding catastrophic forgetting of  pretraining is critical to successful intermediate-task transfer.

The remaining SentEval probing tasks have similar delta values (Figure \ref{fig:transfer1}), which may indicate that there is insufficient variation among transfer performance to derive significant correlations. Among the edge-probing tasks, the more semantic tasks such as coreference (EP-Coref and EP-DPR), semantic proto-role labeling (EP-SPR1 and EP-SPR2), and dependency labeling (EP-Rel) show the highest correlations with our target tasks. As our set of target tasks is also oriented towards semantics and reasoning, this is to be expected.

On the other hand, among the target tasks, we find that ReCoRD, CommonsenseQA and Cosmos QA---all commonsense-oriented tasks---exhibit both high correlations with each other as well as a similar set of correlations with the probing tasks. Similarly, BoolQ, MultiRC, and RTE correlate strongly with each other and have similar patterns of probing-task performance.

\section{Related Work}
\label{sec:rel-work}

Within the paradigm of training large pretrained Transformer language representations via  intermediate-stage training before fine-tuning on a target task, positive  transfer  has been shown in both sequential task-to-task \cite{Phang2018SentenceEO} and  multi-task-to-task \cite{liu-etal-2019-multi, 2019t5} formats. \citet{wang-etal-2019-tell} perform an extensive study on transfer with BERT, finding language modeling and NLI tasks to be among the most beneficial tasks for improving target-task performance. \citet{talmor-berant-2019-multiqa} perform a similar cross-task transfer study on reading comprehension datasets, finding similar positive transfer in most cases, with the biggest gains stemming from a combination of multiple QA datasets. Our work consists of a larger, more diverse, set of intermediate task--target task pairs. We also use probing tasks to shed light on the skills learned by the intermediate tasks.

Among the prior work on predicting transfer performance, \citet{bingel-sogaard-2017-identifying} is the most similar to ours. They do a regression analysis that predicts target-task performance on the basis of various features of the source and target tasks and task pairs. They focus on a multi-task training setting without self-supervised pretraining, as opposed to our single-intermediate task, three-step procedure.

Similar work \citep{lin-etal-2019-choosing} has been done on cross-lingual transfer---the analogous challenge of transferring learned knowledge from a high-resource to a low-resource language. 

Many recent works have attempted to understand the knowledge and linguistic skills BERT learns, for instance by analyzing the language model surprisal for subject--verb agreements \citep{GoldbergBert2019}, identifying specific knowledge or phenomena encapsulated in the representations learned by BERT using probing tasks \citep{tenney2018what,tenney-etal-2019-bert,warstadt-etal-2019-investigating,lin-etal-2019-open,hewitt2019structural,jawahar-etal-2019-bert}, analyzing the attention heads of BERT \citep{clark-etal-2019-bert,Coenen2019,lin-etal-2019-open, phu-syntactic-heads}, and testing the linguistic generalizations of BERT across runs \citep{McCoy-2019-BERT}. However, relatively little work has been done to analyze fine-tuned BERT-style models \citep{wang-etal-2019-tell,warstadt-etal-2019-investigating}.

\section{Conclusion and Future Work}

This paper presents a large-scale study on when and why intermediate-task training works with pretrained models. We perform experiments on RoBERTa with a total of 110 pairs of intermediate and target tasks, and perform an analysis using 25 probing tasks, covering different semantic and syntactic phenomena. Most directly, we observe that tasks like Cosmos QA and HellaSwag, which require complex reasoning and inference, tend to work best as intermediate tasks.

Looking to our probing analysis, intermediate tasks that help RoBERTa improve across the board show the most positive transfer in downstream tasks. However, it is difficult to draw definite conclusions about the specific skills that drive positive transfer. Intermediate-task training may help improve the handling of syntax, but there is little to no correlation between target-task and probing-task performance for these skills. Probes for higher-level semantic abilities tend to have a higher correlation with the target-task performance, but these results are too diffuse to yield more specific conclusions.
Future work in this area would benefit greatly from improvements to both the breadth and depth of available probing tasks. 

We also observe a worryingly high correlation between target-task performance and the two probing tasks which most closely resemble RoBERTa's masked language modeling pretraining objective. Thus, the results of our intermediate-task training analysis may be driven in part by 
\textit{forgetting} of knowledge acquired during pretraining. Our results therefore suggest a need for further work on efficient transfer learning mechanisms.

\section*{Acknowledgments}

This project has benefited from support to SB by Eric and Wendy Schmidt (made by recommendation of the Schmidt Futures program), by Samsung Research (under the project \textit{Improving Deep Learning using Latent Structure}), by Intuit, Inc., and by NVIDIA Corporation (with the  donation of a Titan V GPU).

\bibliography{acl2020}
\bibliographystyle{acl_natbib}

\clearpage

\appendix

\section{Correlation Between Probing and Target Task Performance}

\begin{figure*}[th!]
\centering
  \includegraphics[width=1\textwidth]{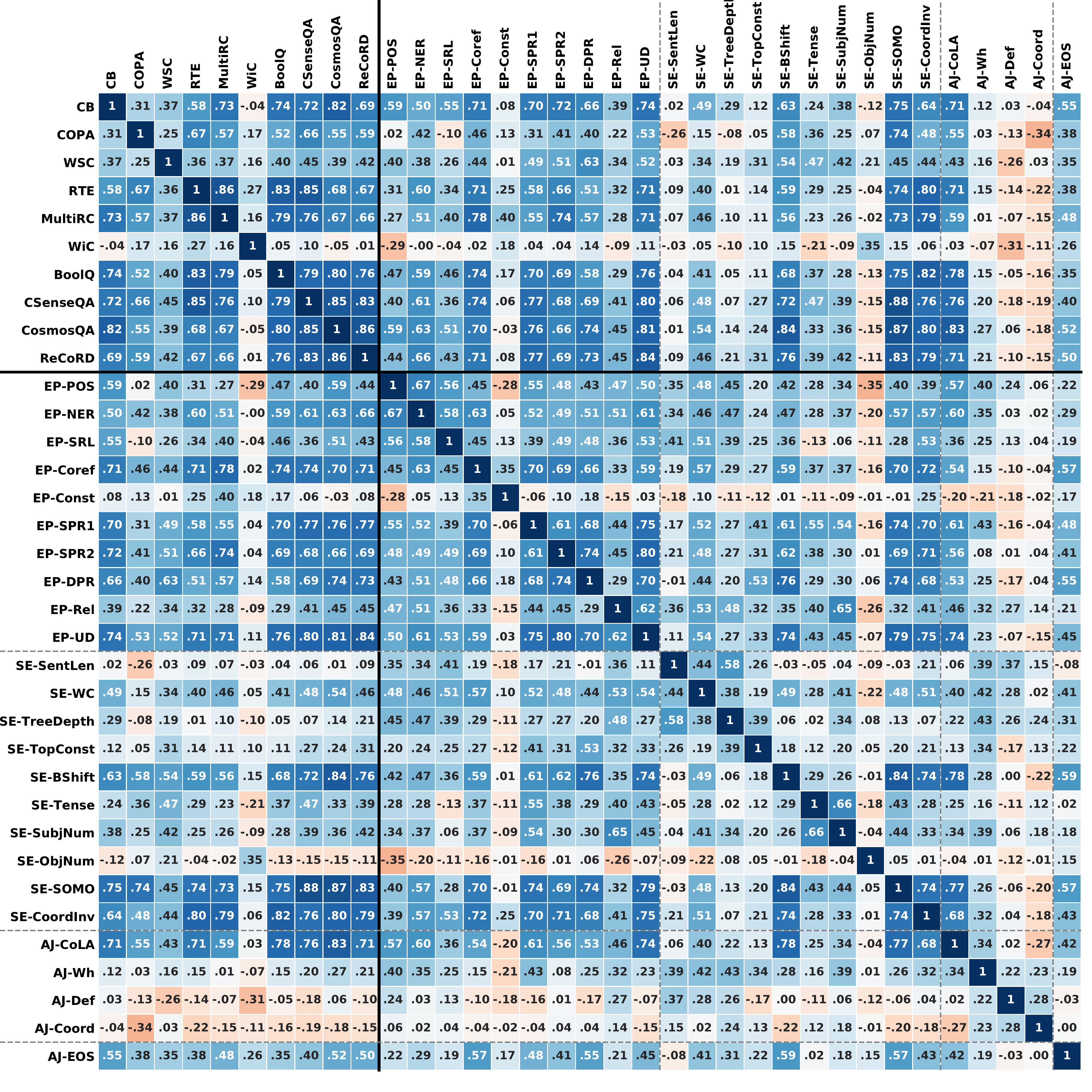}
  \caption{Correlations between probing and target task performances. Each cell contains the Spearman correlation between probing and target tasks performances across training on different intermediate tasks and random restarts.}
  \label{fig:correl_big_spearman}
\end{figure*}

\begin{figure*}[th!]
\centering
  \includegraphics[width=1\textwidth]{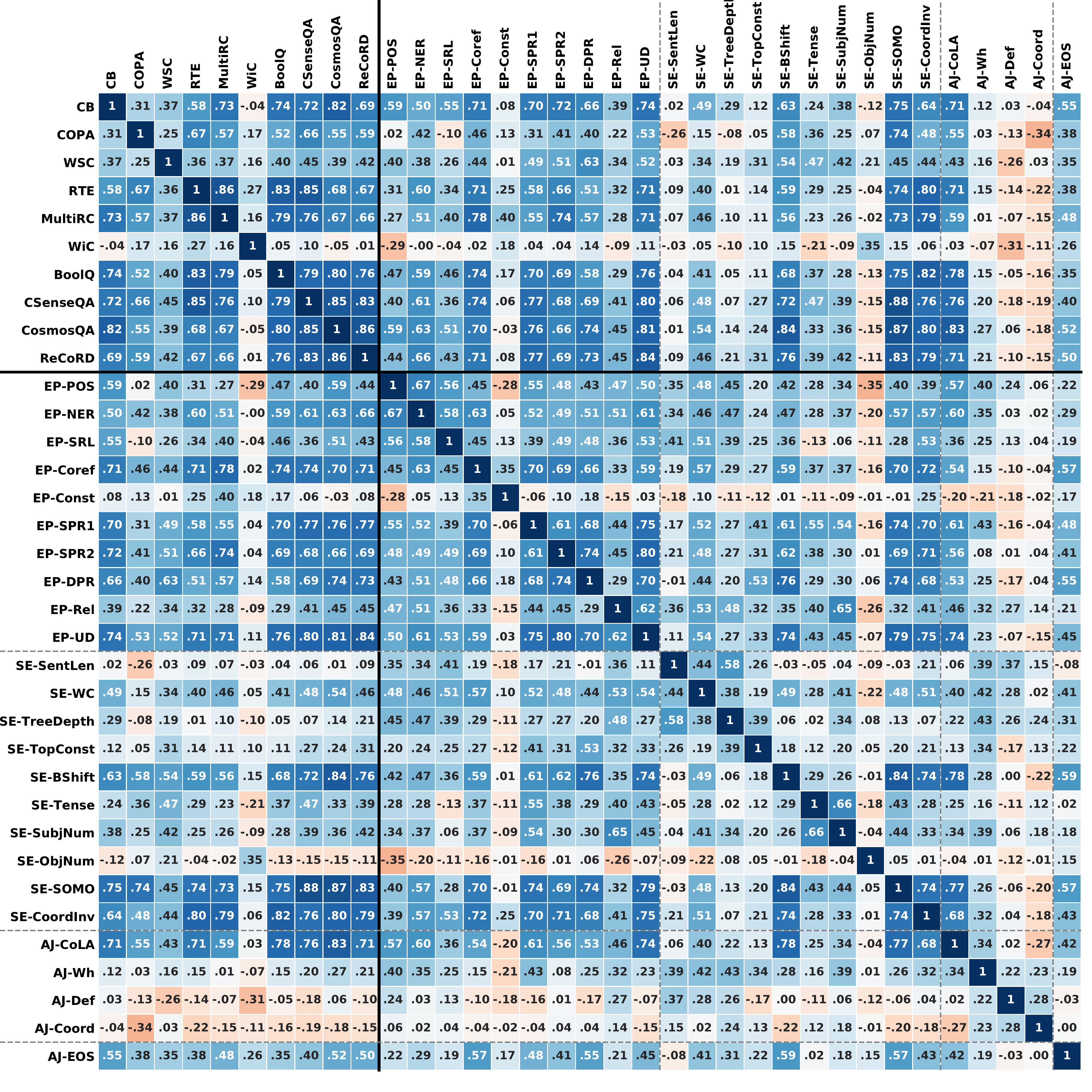}
  \caption{Correlations between probing and target task performances. Each cell contains the Pearson correlation between probing and target tasks performances across training on different intermediate tasks and random restarts.}
  \label{fig:correl_big_pearson}
\end{figure*}

Figure \ref{fig:correl_big_spearman} shows the correlation matrix using Spearman correlation and Figure \ref{fig:correl_big_pearson} shows the matrix using Pearson correlation.

\section{Effect of Intermediate Task Size on Target Task Performance}

\begin{figure*}[th!]
\centering
  \includegraphics[width=1\textwidth]{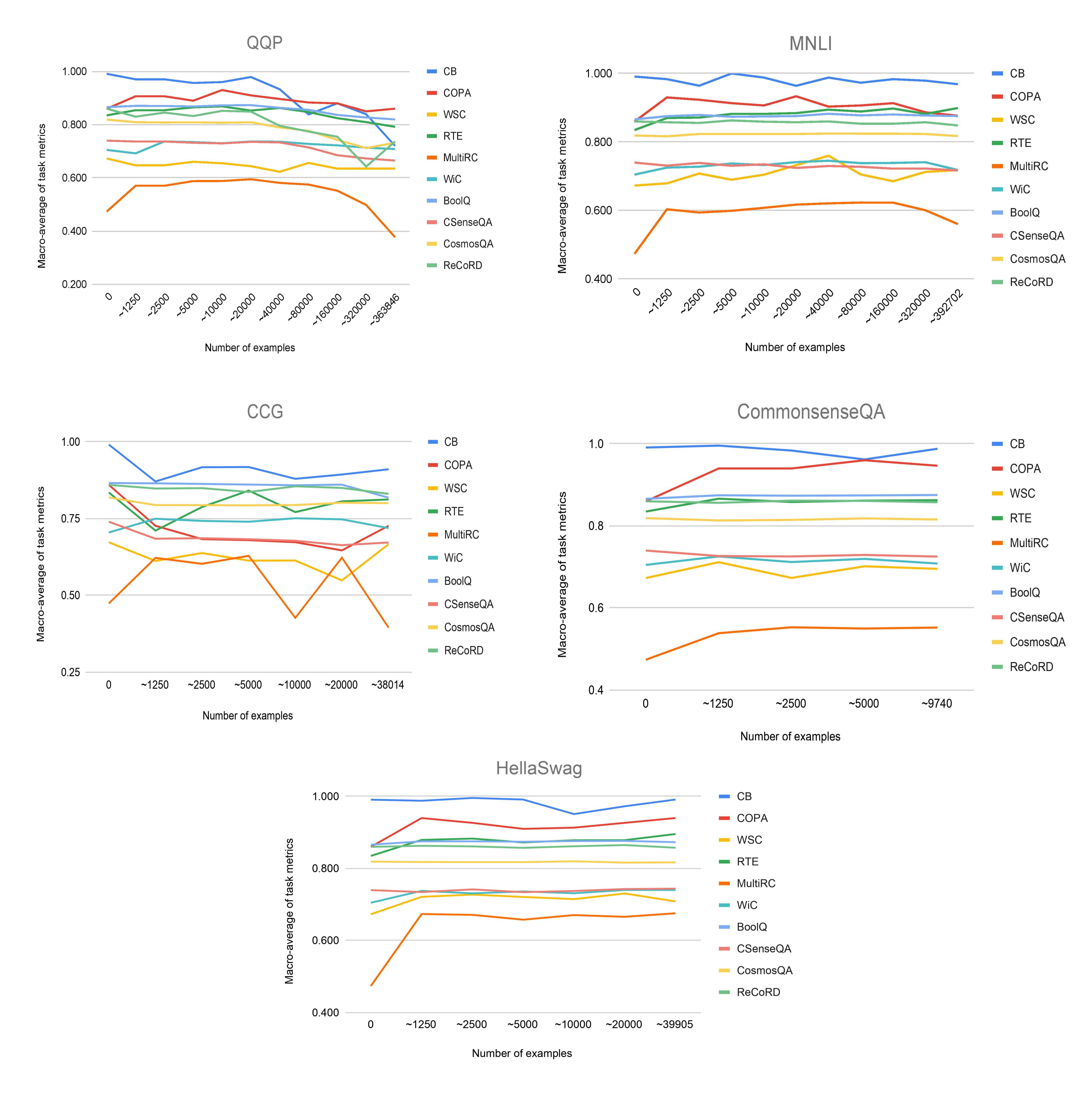}
  \caption{Results of experiments on impact of intermediate task data size on downstream target task performance. For each subfigure, we finetune RoBERTa over a variety of dataset size (sampled randomly from the dataset). We report the macro-average of each target task's performance metrics after finetuning on each dataset size split. }
  \label{fig:frac_experiments}
\end{figure*}

Figure \ref{fig:frac_experiments} shows the effect of dataset size on intermediate task training on downstream target task performance for five intermediate tasks, which were picked to maximize the variety of original intermediate task sizes and effectiveness in transfer learning abilities.

\end{document}